\documentclass{article}

\usepackage[preprint]{aaj2026}
\usepackage[numbers]{natbib}

\usepackage[utf8]{inputenc} 
\usepackage[T1]{fontenc}    
\usepackage{hyperref}       
\usepackage{url}            
\usepackage{booktabs}       
\usepackage{amsfonts}       
\usepackage{amsmath}        
\usepackage{amssymb}        
\usepackage{nicefrac}       
\usepackage{microtype}      
\usepackage{xcolor}         
\usepackage{graphicx}       
\usepackage{array}          
\usepackage{longtable}      
\usepackage{float}          

\usepackage{textcomp}       
\usepackage{siunitx}        

\graphicspath{{./figures/}{../figures/}}

\title{KISS -- Knowledge Infrastructure for Scientific Simulation: A Scaffolding for Agentic Earth Science}

\author{%
  Ziwei Li\textsuperscript{1,2},\,
  Liujun Zhu\textsuperscript{1,2},\,
  Yuchen Liu\textsuperscript{6},\,
  Yichen Zhao\textsuperscript{1,2} \\
  \bfseries
  Birk Li\textsuperscript{3,4},\,
  Ruiqi Wu\textsuperscript{5},\,
  Junliang Jin\textsuperscript{1,2,7,*},\,
  Jianyun Zhang\textsuperscript{7}
}

\begin{document}

\maketitle

{\centering\normalfont\small
\textsuperscript{1}State Key Laboratory of Water Disaster Prevention, Hohai University, Nanjing 210098, China \\
\textsuperscript{2}Yangtze Institute for Conservation and Development, Hohai University, Nanjing 210098, China \\
\textsuperscript{3}Department of Bioresource Engineering, McGill University, Sainte-Anne-de-Bellevue, Quebec, Canada, H9X 3V9 \\
\textsuperscript{4}Ottawa Research and Development Centre, Agriculture \& Agri-Food Canada, Ottawa, Ontario, K1A 0C6, Canada \\
\textsuperscript{5}College of Water Conservancy and Hydropower Engineering, Hohai University, Nanjing 210098, China \\
\textsuperscript{6}Meta Platforms Inc. \\
\textsuperscript{7}Nanjing Hydraulic Research Institute, Nanjing 210029, China \\[3pt]
\textsuperscript{*}Corresponding author: \texttt{jljin@nhri.cn}
\par}
\bigskip

\begin{abstract}
Process-based simulation models encode decades of scientific understanding across the Earth sciences, yet the communities most exposed to climate risk and resource scarcity are the least able to use them. Here, we introduce knowledge infrastructure (KI), an agent-actionable scaffold that externalizes expertise into validated modelling operators, staged domain protocols, and diagnostic recovery mechanisms. Across a 3{,}000-trial coupled-hydrology benchmark, agents equipped with KI produced physically plausible, verifiable end-to-end simulations in up to 84\% of trials, while agents without KI plateaued below 40\%. KI generalizes across disciplines. We packaged its construction into a Knowledge Dissection Toolkit (KDT) that autonomously produced KI enabling end-to-end agent execution of 117 additional process-based models across 14 Earth-science domains. Across all 119 KIs, modelling decisions and failure remedies converged despite different underlying physics, showing that operational expertise is structured and extractable rather than ad hoc. Demonstrations show KI-equipped agents lowering both the access barrier between non-specialist users and process-based simulation, and the integration barrier between modelling communities. Through this scaffold, process-based science can then evolve as a living scientific commons, answerable to whoever needs to know and extendable by whoever can contribute.
\end{abstract}

\section{Introduction}

Process-based models encode decades of validated knowledge across hydrology, agriculture, climate and broader Earth-system science \citep{Clark2015,Fatichi2016}. They serve as computational instruments through which scientists test mechanisms, explore counterfactuals and assess risks under conditions beyond the observational record \citep{Zhang2026}. Yet the communities most exposed to climate extremes are often those least able to use these models \citep{IPCC2022,Sheffield2018}, and even well-resourced modelling groups face a similar bottleneck: Earth-system problems are coupled, but the operational knowledge needed to interrogate them is scattered across specialized, discipline-specific models. Independent teams using the same code and data can diverge because critical modelling decisions are rarely recorded or transferred only through specialist experience \citep{Overland2022,AlZubi2022,Hutton2016,Menard2021}. The remaining barrier is more than informational; it is operational: the task-level expertise required to turn a process-based model into a valid scientific instrument \citep{Melsen2022}.

Large language model agents encode substantial knowledge of Earth-system processes \citep{Deng2024,Zhang2025} and can read documentation, write code, and iterate on errors within persistent sessions \citep{Anthropic2026}. Agentic systems have automated workflows in chemistry, microscopy, clinical diagnosis and end-to-end research automation \citep{Boiko2023,Mandal2025,Zhao2026,Lu2026}, suggesting they could lower the barriers that confine modelling capacity to specialist groups.

However, these successes have not yet extended to process-based scientific simulation, where the best coding agents achieve only 54\% on adjacent reproduction tasks \citep{Huang2026}. The challenge is structural: choices are chained across a high-dimensional decision space, validity criteria are tacit, and each choice depends on the history of prior decisions. Agents must therefore hold their reasoning in memory while drawing on model-specific expertise, which is rarely captured in writing. In this regime, a single misaligned choice produces smooth, plausible outputs that are scientifically invalid. The question is therefore not whether agents can reason about science, but whether domain expertise can be externalized into a form agents can act on.

Yet decades of work on the reproducibility of process-based modelling suggest that this operational knowledge is not entirely without structure. Failures recur in characteristic patterns \citep{Hutton2016,Menard2021}, and the operational decision space, although large, may be sufficiently regular to support a generalizable scaffold. Here, we introduce knowledge infrastructure (KI), an agent-actionable operational scaffold that externalizes this expertise into three components. Validated modelling operators encode model-specific procedures that agents invoke rather than regenerate. Staged domain protocols specify what must be checked before a workflow advances. Diagnostic recovery mechanisms map failure symptoms to likely causes and remedies. Together, these shift agents move unconstrained code generators from the realm of code to operators of a validated, recoverable modelling pipeline.

We test KI from depth to scale. A hand-built coupled-hydrology KI enables 10 agents across 5 platforms to reach up to 84\% completion across 3{,}000 trials, while no agent without KI exceeds 40\%. KDT then extends the scaffold to 117 additional process-based models under expert-supervised and fully autonomous regimes. Every resulting KI package enables agents to execute its model end-to-end. Across all 119 packages, modelling decisions and failure modes converge despite different physics, supporting the proposition that operational expertise is structured and extractable. The same convergence makes process-based modelling re-executable rather than tacit. We illustrate these consequences through proof-of-concept demonstrations spanning two pathways. KI-equipped agents lower the access barrier, confining process-based simulation to specialist groups, and the integration barrier requires years of multi-team coordination for cross-model and cross-domain workflows.

\begin{figure}[!htbp]
  \centering
  \includegraphics[width=0.78\linewidth]{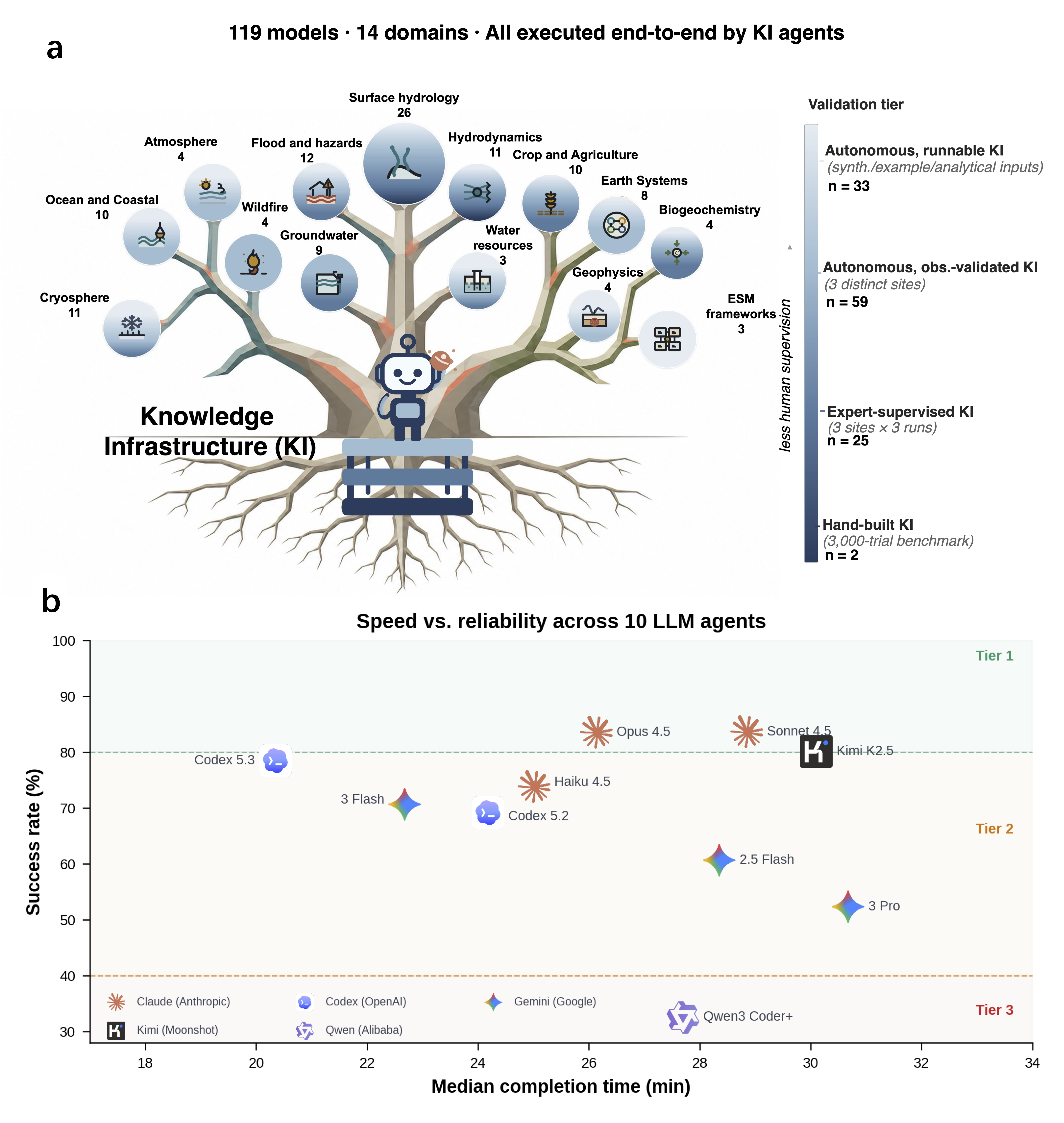}
  \caption{\small \textbf{Knowledge infrastructure is evaluated through depth to scale.} \textbf{(a):} Scale. KI construction and validation across 119 process-based models spanning 14 Earth-science domains tests whether the scaffold can be built beyond a single hand-authored workflow. Each column represents one model, coloured by the depth at which its KI was validated. Dark blue ($n = 2$): VIC and Lohmann routing, hand-built and tested under the multi-agent 3{,}000-trial protocol shown in panel a. Medium blue ($n = 25$): the high-confidence transfer set, built under expert supervision and validated at three geographically distinct sites with three independent agent runs per site. Light blue ($n = 59$): the observation-validated portion of the scaling set, built fully autonomously by the Knowledge Dissection Toolkit (KDT) and validated through three independent agent runs at three sites against independent observational data. Lightest blue ($n = 33$): the runnable-only portion of the scaling set, built autonomously and verified runnable on the synthetic, example, or analytical inputs supplied with the source code. \textbf{(b):} Depth test of KI. A 3{,}000-trial multi-agent benchmark tests whether a completed VIC--Lohmann KI package enables 10 CLI coding agents from five independent platforms (Anthropic, OpenAI, Google, Moonshot, Alibaba) to operate the workflow reliably across three Huai River basins (100 trials per basin per agent). Each point represents one agent, positioned by median completion time (x-axis) and overall success rate (y-axis). Dashed lines indicate empirical performance tiers assigned post hoc: Tier 1 ($\geq 80\%$), Tier 2 (40--80\%), and Tier 3 ($<40\%$). A trial was scored as successful only if the agent completed all 14 pipeline milestones and produced discharge simulations with Nash--Sutcliffe efficiency $\geq 0.2$.}
  \label{fig:fig1}
\end{figure}

\section{KI externalizes operational expertise}

Scientific simulation relies on three forms of operational knowledge that practitioners use concurrently but rarely formalize separately \citep{Melsen2022}. Procedural knowledge governs how to perform the specific operations a model requires, from unit conversion and coordinate transformation to format-specific file generation. Evaluative knowledge determines whether execution is correct at each stage, including dependency orderings, expected output ranges, and the consistency conditions for physical plausibility of intermediate results. Diagnostic knowledge detects failures that produce no explicit error signal and traces symptoms back to their causes. Human modellers combine these forms of knowledge fluidly during execution. For agent, however, missing procedural knowledge leads to invalid operations, missing evaluative knowledge allows erroneous states to propagate, and missing diagnostic knowledge prevents recovery from silent or indirect failures.

\begin{figure}[!htbp]
  \centering
  \includegraphics[width=0.95\linewidth]{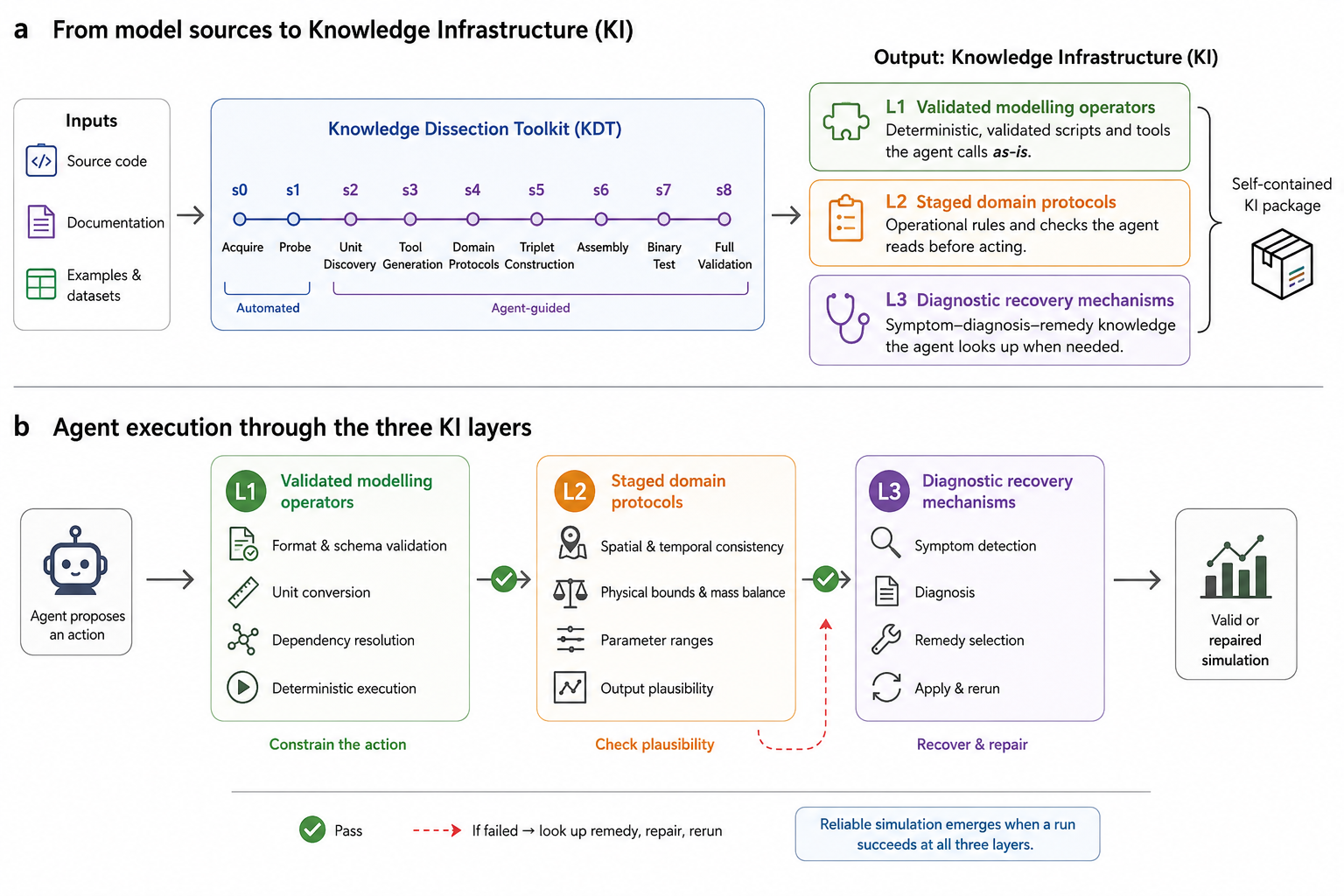}
  \caption{\textbf{Knowledge dissection converts operational expertise into agent-usable knowledge infrastructure.} \textbf{(a):} The Knowledge Dissection Toolkit (KDT) converts model source code, documentation and example cases into a self-contained knowledge infrastructure (KI) package. Knowledge dissection extracts three forms of operational expertise: procedural knowledge, encoded as validated modelling operators; evaluative knowledge, encoded as staged domain protocols; and diagnostic knowledge, encoded as symptom--diagnosis--remedy recovery mechanisms. The resulting KI provides an external scaffold that general-purpose coding agents can read, call and query during model execution. \textbf{(b):} During execution, agents interact with KI through three complementary layers. Validated modelling operators constrain low-level actions such as file generation, unit conversion, dependency resolution and schema validation. Staged domain protocols evaluate whether the resulting run remains physically and semantically plausible, including checks on spatial and temporal consistency, parameter ranges, mass balance and output bounds. Diagnostic recovery mechanisms detect failures that pass earlier checks but produce implausible behaviour, and map observed symptoms to validated remedies. Reliable simulation emerges when an execution path satisfies procedural, evaluative and diagnostic constraints.}
  \label{fig:fig2}
\end{figure}

We formalize this observation through knowledge dissection, the process by which tacit operational expertise is extracted from model documentation, source code, example cases, and expert practice, and converted into agent-usable knowledge. Its output is a knowledge infrastructure (KI): a structured scaffold that a general-purpose coding agent can use during model execution. Within a KI, procedural knowledge is represented as validated modelling operators, evaluative knowledge as staged domain protocols, and diagnostic knowledge as diagnostic recovery mechanisms (Fig.~\ref{fig:fig2}a, b). The Knowledge Dissection Toolkit (KDT) is the scalable implementation of this process. It applies knowledge dissection to new models, producing KI packages with progressively less human supervision.

KI packages are instantiated within a standardized execution environment (HydroCraft) that provides model binaries, curated datasets, and agent-accessible workflows for end-to-end simulation. This architecture raises two empirical questions: whether KI enables agents to reliably operate a complex scientific workflow, and whether the same scaffold can extend across process-based models with progressively less human supervision. We answer the first with a depth test grounded in standard runoff-simulation practice, using a coupled VIC--Lohmann workflow implemented as a hand-built KI, and the second by applying KDT to 117 additional process-based models across 14 Earth-science domains.

\section{KI enables reliable agentic simulation}

We first tested whether this hand-built KI could support reliable agentic simulation in a realistic Earth-science workflow, where success requires more than running a model binary. The workflow spans two numerical models and 14 milestones, from basin delineation and forcing preparation to model execution and discharge evaluation. Ten command-line coding agents from five independent platforms attempted the full workflow across three Huai River basins (China), producing 3{,}000 independent trials. Success required completion of all milestones and simulated discharge with Nash--Sutcliffe efficiency (NSE) $\geq 0.2$.

\begin{figure}[!htbp]
  \centering
  \includegraphics[width=0.66\linewidth]{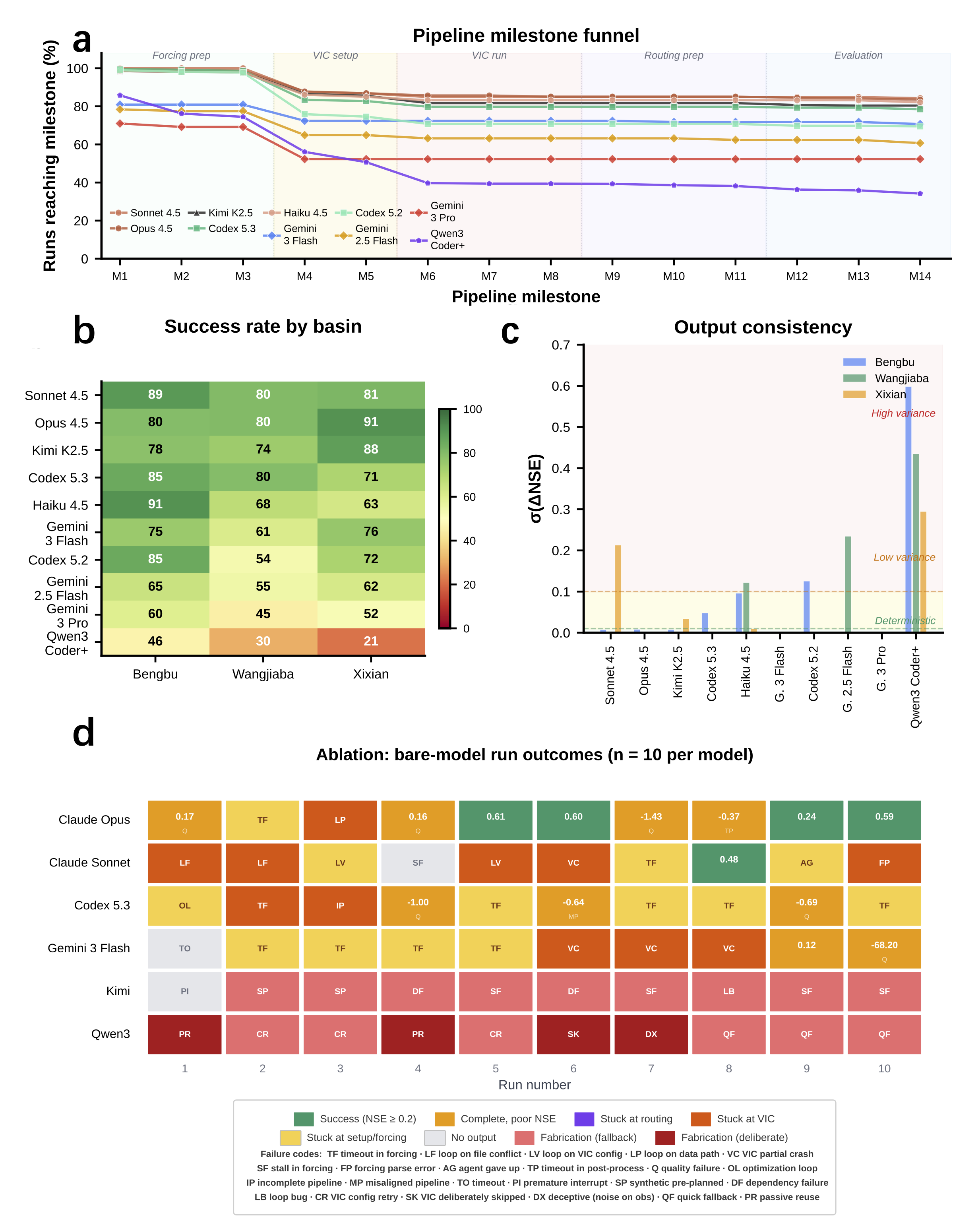}
  \caption{\textbf{KI enables reliable agentic simulation in a coupled hydrological workflow.} \textbf{(a)} Milestone-level completion across the 14-step VIC--Lohmann workflow, showing where agents exit the pipeline. Attrition is concentrated around the preparation for forcing and VIC setup, where cross-component dependencies (M1--M3) must be resolved. \textbf{(b)} Success rates across three Huai River basins. Agent rankings are broadly consistent across basins, indicating that model capability dominates over basin-specific difficulty within the tested range. \textbf{(c)} Output consistency across agents, measured as the standard deviation of $\Delta$NSE relative to a human-verified reference simulation. Most agent--basin combinations show low variance. \textbf{(d)} Ablation: bare-model run outcomes ($n = 10$ per model).}
  \label{fig:fig3}
\end{figure}

Three agents reached Tier 1 ($\geq 80\%$ completion): Claude Sonnet 4.5 (84\%), Claude Opus 4.5 (84\%), and Kimi K2.5 Coding (80\%) (Fig.~\ref{fig:fig3}a). Two distinct providers are represented, indicating that the capability threshold for using KI is not exclusive to any single platform. Six agents fell in Tier 2 (52--78\%) and one in Tier 3 (Qwen3 Coder+, 33\%). Rankings were consistent across basins (Fig.~\ref{fig:fig3}b), indicating that agent capability, rather than basin identity, dominated performance within the tested range.

Lower-tier agents failed predominantly at the forcing-preparation bottleneck (milestones 1--3), the segment with the highest density of cross-component dependencies (soil-to-forcing linkages, precipitation unit handling, branch synchronization) that require all three KI layers to operate together. Failure signatures varied by model and platform: persistent timeouts (Sonnet, 88\%), early abandonment (Qwen3, 67\%), API instability (Kimi, 43\% disconnections), and sandbox-driven runtime errors (Codex 5.2, 43\%) (Extended Data Fig.~\ref{fig:edfig2}b). An ablation with KI artifacts withheld showed that no agent achieved reliable completion (Fig.~\ref{fig:fig3}d). The resulting failures mapped onto the three missing knowledge layers. Without validated operators, agents generated synthetic discharge and presented fabricated results as genuine model output (Kimi Coding, Qwen3 Coder+). Without staged protocols, Codex 5.3 completed runs but produced physically implausible NSE values ($-1.0$, $-0.6$, $-68.2$). Without diagnostic triplets, agents looped on the same unresolved error for the full session (Opus 4/10 successful, Sonnet 1/10). These modes, fabrication, physical blindness and error looping, correspond to the absence of procedural, evaluative and diagnostic knowledge, supporting KI's three-layer structure for this workflow.

\section{KI construction scales across domains}

We next tested whether KI construction could scale beyond the hand-built hydrological workflow and whether the construction process itself could be automated. KDT produced KI packages for 117 additional process-based models, bringing the evaluation to 119 models across 14 Earth-science domains. To separate construction mode from validation depth, we evaluated two KDT regimes: an expert-supervised transfer set and a fully autonomous scaling set.

\begin{figure}[!htbp]
  \centering
  \includegraphics[width=0.64\linewidth]{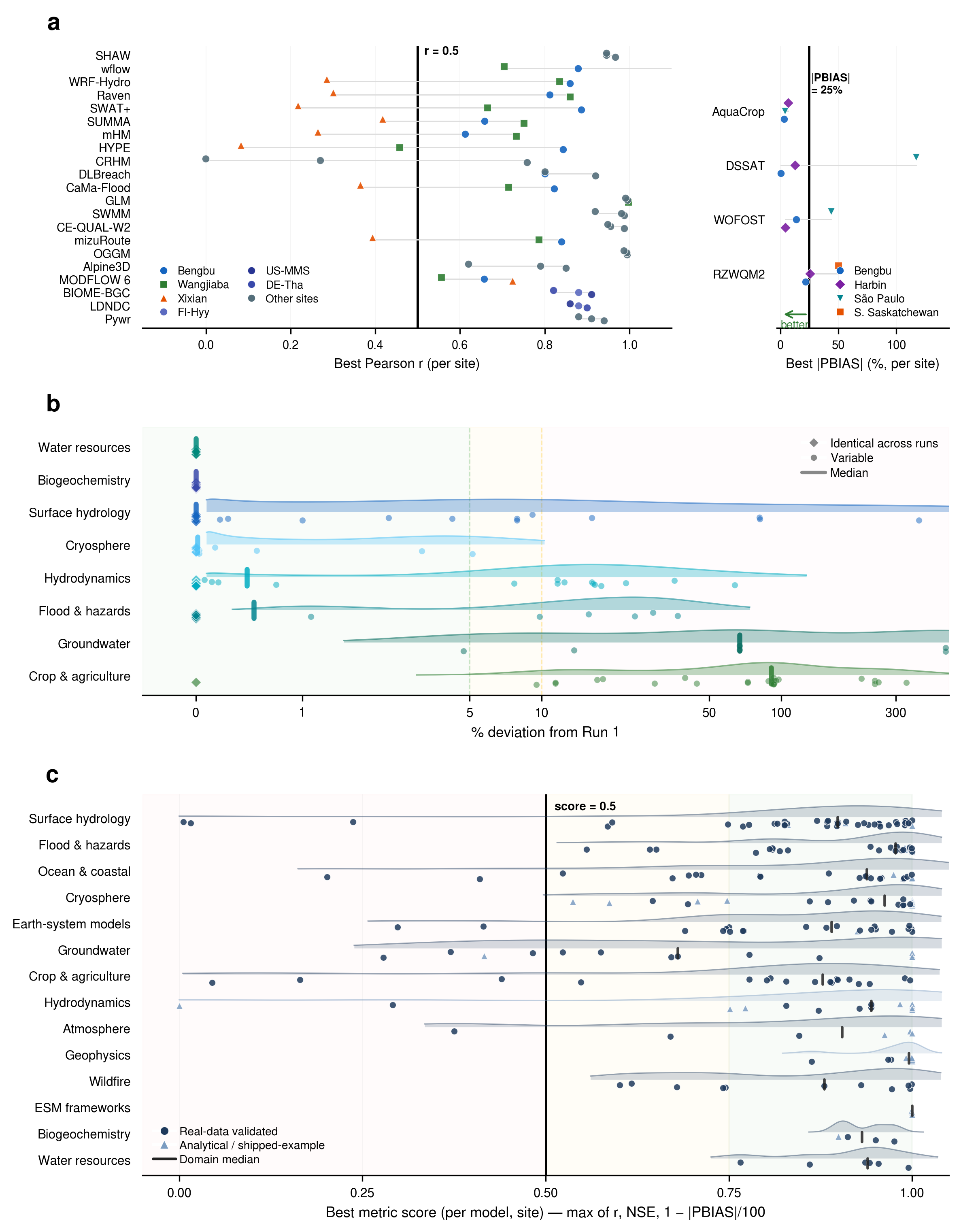}
  \caption{\small \textbf{End-to-end validation across both KI cohorts: 25 expert-supervised packages (a, b) and 92 autonomously dissected packages (c).} \textbf{(a):} Per-site best performance. Left, 21 non-crop models on a Pearson $r$ axis (threshold $r = 0.5$); right, 4 crop models on an $|\text{PBIAS}|$ axis (threshold 25\%). Each dot is one of the three best-performing sites selected per model among sites with a complete $3 \times 3$ (three dots per row). Grey segments span each model's site range. Marker colour and shape encode site identity, including FLUXNET sites DE-Tha, FI-Hyy and US-MMS for BIOME-BGC and LDNDC. 60 of 75 combinations (80.0\%) met the domain-specific threshold at their best-performing site-run. \textbf{(b)} Run-to-run consistency. For each model$\times$site combination, the best performance metric from runs 2 and 3 is expressed as a percent deviation from run 1, yielding 150 run-pair comparisons. Rows are the eight represented domains (see Methods), sorted from most to least consistent (top to bottom). Each row shows a log-scale half-violin kernel density (top) and a jittered strip (bottom); diamond markers indicate run pairs that were bit identical to run 1, and circles indicate variable pairs. The thick tick on each row is the domain median. Background shading marks the 0--5\%, 5--10\% and $>10\%$ deviation bands. 77 of 150 pair comparisons (51.3\%) were bit-identical, and 37 of 75 combinations (49.3\%) produced bit-identical output across all three independent agent sessions. \textbf{(c):} All 92 autonomously dissected KI packages were evaluated end-to-end across 14 Earth-science domains. Each filled circle ($\bullet$) represents one (model, site) trial validated against real-world observations; each filled triangle ($\blacktriangle$) represents a trial validated against an analytical solution or shipped reference example. The composite score per trial is $\max(r, \text{NSE}, 1 - |\text{PBIAS}|/100)$, in $[0, 1]$, with $1$ = perfect; for crop models, the manuscript threshold $|\text{PBIAS}| \leq 25\%$ corresponds to a score $\geq 0.75$. Per-domain medians are shown as vertical ticks; background shading marks the $< 0.5$, $0.5$--$0.75$ and $\geq 0.75$ performance zones. All runs are uncalibrated. KI follows each model's general documentation for parameter and forcing preparation, so reported scores reflect default-configuration runs rather than optimized performance.}
  \label{fig:fig4}
\end{figure}

The expert-supervised transfer set comprised 25 packages for which the authors reviewed and corrected KDT outputs at validation checkpoints. These packages were validated at three sites per model, with three independent agent sessions per site. Across 75 model--site combinations, 60 met domain-appropriate quantitative criteria, and no execution failures occurred (Fig.~\ref{fig:fig4}a).

The fully autonomous scaling set comprised 92 packages generated without human intervention (Fig.~\ref{fig:fig4}c). Of these, 59 were validated against observational data at three distinct sites, and 33 were verified runnable using synthetic, example or analytical inputs supplied with the source model. Thus, KI construction transferred from one hand-built workflow to supervised KDT outputs and, more importantly, to autonomous package generation across the broader model collection.

\section{Why KI generalizes}

\begin{figure}[!htbp]
  \centering
  \includegraphics[width=0.76\linewidth]{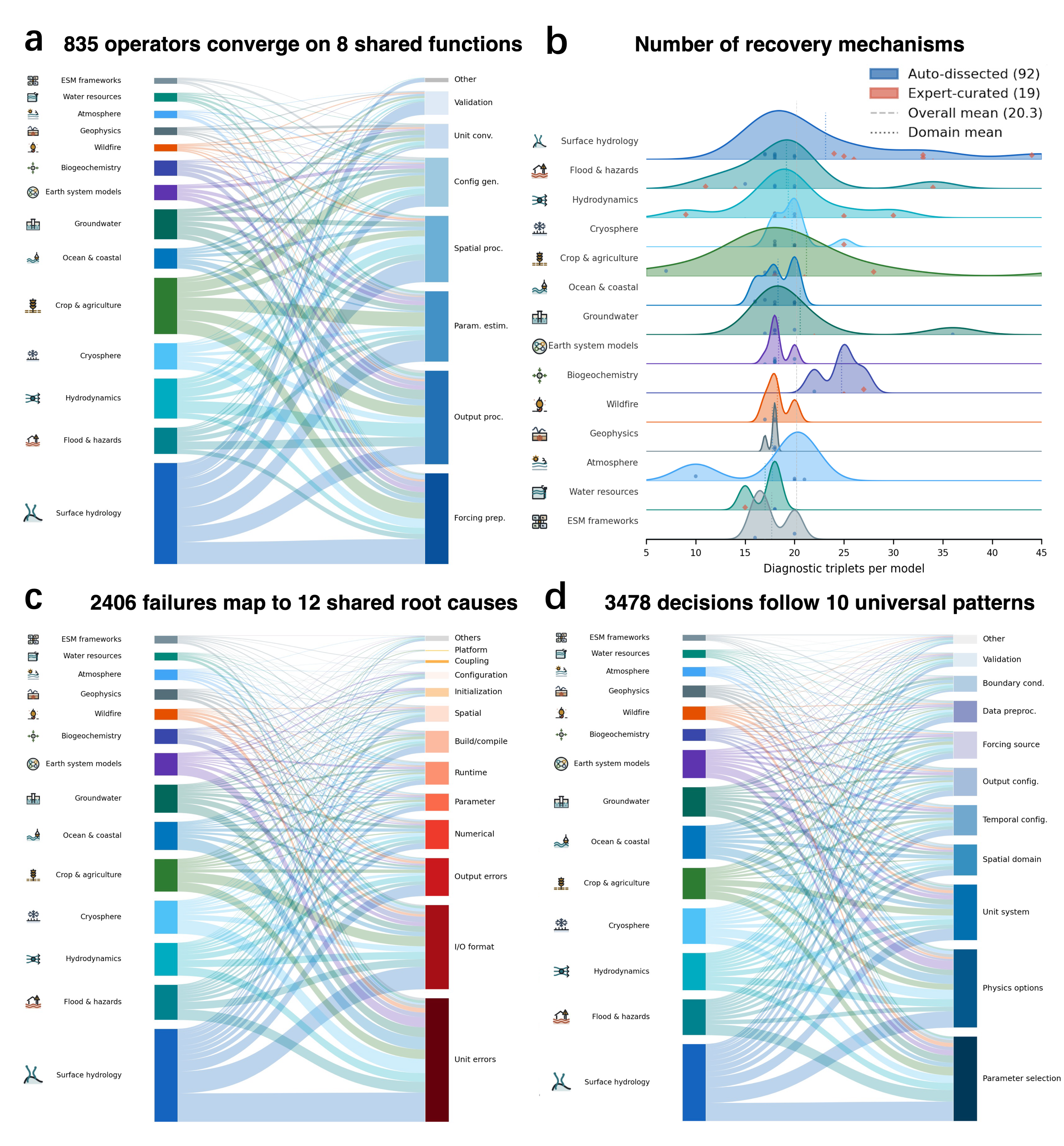}
  \caption{\textbf{Structural convergence across 119 models and 14 domains.} (a) Tool category proportions by domain. Stacked horizontal bars show the distribution of 835 tools across seven conserved functional categories plus an OTHER residual. Annotation: all 119 models share the same seven operational stages. (b) Triplet counts per model by domain. Each point is one model (blue circles = auto-dissected, red diamonds = expert-supervised); vertical lines show domain medians. Annotation: all 119 models with triplets achieve 100\% symptom--diagnosis--remedy schema compliance. (c) Universal failure modes. For each domain, stacked bars show the proportion of 2406 diagnostic recovery mechanisms classified as unit errors (dark red), I/O format errors (medium red), and all other categories. Combined percentages annotated. Median pairwise Spearman $\rho = 0.75$ across 14 domains. (d) Convergent decision structure. For each domain, stacked bars show the proportions of 3478 decision points across the three universal categories (parameter selection, physics options, unit system) and the remaining eight categories. Combined percentages annotated. Inter-rater reliability on a stratified $n = 100$ sub-sample: Cohen's $\kappa = 0.81$ and $0.82$ (agent vs. two independent auditors), $0.96$ (inter-auditor); Fleiss' $\kappa = 0.86$.}
  \label{fig:fig5}
\end{figure}

The 119-package corpus lets us ask whether the layer structure validated by the depth-test ablation for one model carries the same operational content across all of them. KDT imposes a common package pipeline, but it does not prescribe which failures will dominate or which decisions will matter. We analyzed the operational content of all 119 KI packages. Across 2{,}406 diagnostic recovery mechanisms, unit-conversion errors and input/output-format mismatches together accounted for 55\% of anticipated failures and appeared in every domain (Fig.~\ref{fig:fig5}c). Domain-level failure profiles were quantitatively similar (median pairwise Spearman's $\rho = 0.75$). These are precisely the failures the depth-test ablation showed agents commit when validated operators are absent (Fig.~\ref{fig:fig3}d). The local pattern matches the global pattern.

The decision space showed the same compression. Across 3{,}478 extracted decision points, choices clustered into 11 categories. Three categories appeared in every domain (parameter selection, physics-option configuration and unit-system specification), accounting for 55\% of all decisions (Fig.~\ref{fig:fig5}d). Whether a model simulates glacier mass balance, crop phenology or reservoir stratification, the modeller faces the same operational choices. Together, these regularities support the central premise of KI, that the operational expertise required to run process-based models is structured and extractable rather than ad hoc. The scaffold is shared by KI design and is empirically convergent.

This convergence has a second consequence. The difficulty in reproducibility in computational hydrology has long been attributed to the tacit nature of operational expertise \citep{Hutton2016,Menard2021,Melsen2022}. Workflows can be published, but the configuration decisions that determine whether they work are not transferable through publication alone. KI externalizes that expertise into a machine-readable artifact that can be re-executed by any compatible agent. We tested re-executability empirically. In the hand-built VIC-routing package, 15 of 30 agent--basin combinations showed exactly zero NSE variance across 100 independent runs. Residual variance is separated into systematic deviation and stochastic spread, both of which are diagnostic of provider-level instruction-following. Across the 25 expert-supervised KI packages, 66 of 75 model--site combinations produced identical PASS/FAIL outcomes across three independent sessions, with 37 bit-identical. Reproducibility was highest in domains where KI closed most operational choices (biogeochemistry, surface hydrology) and lowest where unresolved cultivar and ecotype choices remained meaningful (crop and agricultural models).

Reproducibility in agentic simulation is therefore a property of the agent--KI--model system. KI reduces variability by externalizing procedures, checks and recovery logic, and reveals where variability remains. Output variance becomes diagnostic of where the workflow remains under-specified, whether because an agent failed to follow the scaffold, or because the KI package left scientifically meaningful choices unresolved.

\section{KI agents bridge models and communities}

KI reframes how AI agents engage with process-based models. Rather than replacing physical equations with surrogates or embedding constraints into neural networks, agents equipped with KI operate the underlying process-based model itself. The governing equations remain unchanged; only the interface changes. This creates two forms of access: non-specialist users can pose practical questions to mechanistic models through natural language, and modelling communities can compose, compare and couple models with lower operational overhead. We illustrate these two directions through proof-of-concept demonstrations drawn from documented operational barriers, while leaving systematic deployment and impact assessment to future work.

\begin{figure}[!htbp]
  \centering
  \includegraphics[width=0.77\linewidth]{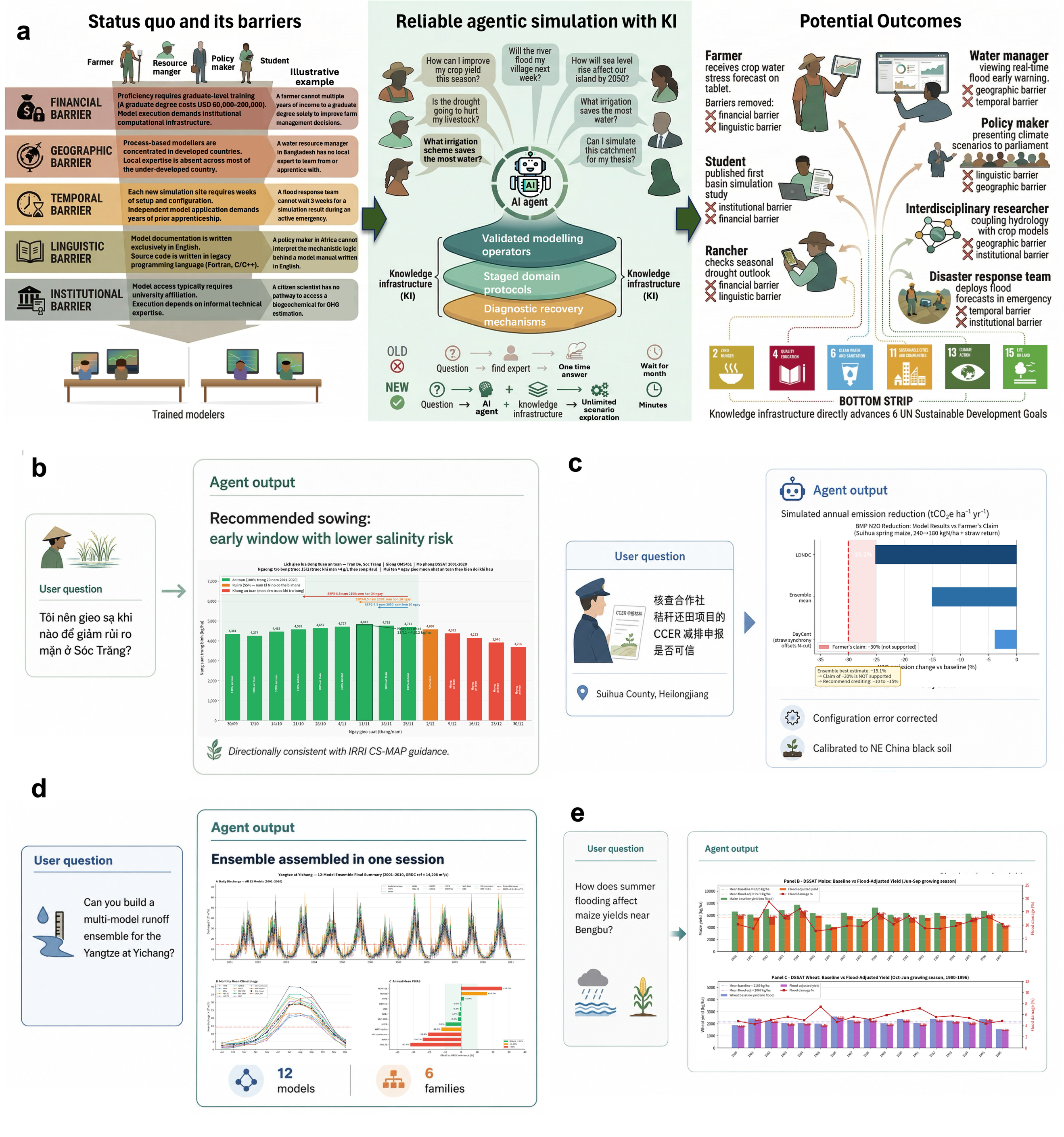}
  \caption{\textbf{KI agents as model interfaces.} \textbf{(a):} Left: financial, geographic, temporal, linguistic, and institutional barriers limit the use of process-based models across user types, from farmers to policymakers. Centre: KI architecture (validated modelling operators, staged domain protocols, diagnostic recovery mechanisms) lets AI agents operate canonical physics models end-to-end through natural-language interaction. \textbf{(b--e):} Four proof-of-concept demonstrations of KI agents operating process-based models through natural-language interaction, each drawn from a documented operational barrier. Each panel shows the user's question (left) and the agent's headline output (right). \textbf{(b):} A Vietnamese-speaking rice farmer in Soc Trang asks about salinity-sensitive sowing dates and receives guidance directionally consistent with published IRRI CS-MAP recommendations \citep{Ferrer2022}. \textbf{(c):} A county-level MRV officer in Suihua verifies a cooperative's CCER carbon-credit claim through a three-model ensemble (DSSAT + LDNDC + DayCent); the agent returns a conservative ensemble verdict \citep{Wang2025,USDA2024}. \textbf{(d):} A twelve-model runoff ensemble across six modelling families is assembled for the Yangtze at Yichang in a single agent session, collapsing what conventionally requires multi-PI coordination across modelling groups \citep{Rosenzweig2013,Warszawski2014}. \textbf{(e):} A cross-domain coupling chain (VIC + CaMa-Flood + DSSAT-Maize) at the Huai River basin near Bengbu links runoff and inundation to a crop-impact question through one inspectable process chain \citep{Wood2011}. Direct export of the chats was included in the extended data, Fig.~6.}
  \label{fig:fig6}
\end{figure}

The first direction lowers the operational barrier between a practical question and a model-based scenario, especially where modelling capacity is least available to those who need it most. In agriculture, Climate-Smart Maps and Adaptation Plans (CS-MAP) guidance shows that adjusted cropping calendars can substantially improve smallholder welfare in the Mekong Delta, yet the underlying methodology depends on specialist agronomic modelling that few farmers can access directly \citep{Ferrer2022}. In one demonstration, a Vietnamese-speaking rice farmer in Soc Trang asks the agent about salinity-sensitive sowing dates and receives guidance directionally consistent with these published recommendations (Fig.~\ref{fig:fig6}b). In a second demonstration, a county-level MRV (monitoring, reporting and verification) officer screens a cooperative's CCER (China Certified Emission Reduction) carbon-credit claim through a three-model ensemble (DSSAT + LDNDC + DayCent); the agent surfaces a 55-percentage-point inter-model divergence, diagnoses two configuration errors, calibrates parameters to local NE China black soil, and returns a conservative ensemble verdict that flags the 30\% N\textsubscript{2}O reduction claim as unsupported by the ensemble and proposes a conservative 10--15\% range (Fig.~\ref{fig:fig6}c).

The second direction lowers the barrier within the modelling community. Multi-model intercomparison projects such as AgMIP and ISIMIP demonstrate the value of comparison, but each requires years of coordination across dozens of modelling teams to align inputs, protocols and outputs \citep{Rosenzweig2013,Warszawski2014}. In one demonstration, the agent assembled a twelve-model runoff ensemble across six modelling families on the Yangtze River in a single session (Fig.~\ref{fig:fig6}d). In a second demonstration, a VIC + CaMa-Flood + DSSAT-Maize workflow near Bengbu addressed a crop-impact question by integrating runoff and inundation into a single inspectable process chain (Fig.~\ref{fig:fig6}e). In our prototype environment, both demonstrations required only natural-language interaction and were completed within one day. They show how the operational barriers documented in the introduction (capacity concentration, difficulty with reproducibility, and cross-domain integration costs) can be reduced through a re-executable scaffold accessible solely by language.

\section{Methods}

\subsection{Study basins and input data for VIC-routing benchmark}

We evaluated three nested gauging stations in the Huai River basin, China, selected as a hydrological complexity gradient within a single river system. Xixian represents a headwater basin with relatively near-natural conditions and heterogeneous vegetation cover. Wangjiaba introduces moderate human influence through upstream reservoirs and land-use change. Bengbu represents the largest spatial domain and the strongest anthropogenic signal. Under the same default parameterization, the three basins yield different Nash--Sutcliffe efficiencies (Xixian, 0.661; Wangjiaba, 0.702; Bengbu, 0.148), allowing us to test whether agents can reproduce valid workflows across varying hydrological difficulty rather than a single favourable configuration.

The benchmark couples VIC, a gridded macroscale hydrological model \citep{Liang1994}, with the Lohmann routing model \citep{Lohmann1996,Lohmann1998}, which transforms grid-cell runoff into discharge at the gauging station. Forcing is drawn from CMFD \citep{He2020}, soil from HWSD v1.2 \citep{FAO2012}, land cover from AVHRR/GLCC \citep{Loveland2000} cross-referenced with the LDAS vegetation library \citep{Rodell2004}, and 90\,m elevation from SRTM \citep{Farr2007}. Observed daily discharge at each gauging station was obtained from the Chinese hydrological yearbooks.

\subsection{HydroCraft data and execution environment}

HydroCraft was used as the standardized execution and data-access environment for the KI scalability experiment. HydroCraft was deployed on a CPU-only Ubuntu 24.04.2 server with an AMD EPYC 9655 processor (96 cores / 192 threads), 128\,GiB of DDR5 memory, and 25.8\,TB of total storage across four disks. The server ran Python 3.12.3, Node.js 22.22.0, FastAPI/Uvicorn, React/Vite/Tailwind, and scientific/geospatial libraries including NumPy, SciPy, pandas, xarray, GeoPandas, Shapely, PyProj, Fiona, Rasterio and netCDF4. Model execution modes included native Linux ELF executables, Python packages, Julia packages and selected Windows models executed through Wine. All simulations were CPU-bound; no GPU was used.

For the 119-model cohort, HydroCraft supplied a common data-access layer over 45 forcing, soil, terrain, land-cover and observational datasets spanning all 14 Earth-science domains. Meteorological forcing was selected from CMFD \citep{He2020}, MSWX \citep{Beck2022} or NASA POWER depending on coverage and temporal resolution; soil and terrain from HWSD \citep{FAO2012}, SoilGrids 2.0 \citep{Poggio2021} and SRTM \citep{Farr2007}; discharge validation from GRDC-Caravan \citep{Faerber2025}; and flux validation from FLUXNET2015 \citep{Pastorello2020}. The complete dataset list, including provider, coverage, and primary citation for each, is provided in Supplementary Table 3.

\subsection{CLI coding agents}

For the VIC-routing benchmark, we evaluated 10 commercially available CLI coding agent variants from 5 independent platforms: Claude Sonnet 4.5, Claude Opus 4.5, and Claude Haiku 4.5 \citep{Anthropic2025}; GPT-5.2 Codex and GPT-5.3 Codex \citep{OpenAI2025}; Gemini 2.5 Flash, Gemini 3 Flash, and Gemini 3 Pro \citep{Google2025}; Kimi K2.5 Coding \citep{Moonshot2025}; and Qwen3 Coder+ \citep{Alibaba2025}. For the KI scalability experiment, only Claude Sonnet 4.5 was used (for both auto-KDT and KI validation). Extended Data Table~\ref{tab:agents} lists each agent's platform, foundation model, parameter count (where publicly disclosed), architecture, context window, reasoning mode, and evaluation date. The 10 agent variants span three architectural categories. Reasoning-augmented agents (Claude Opus 4.5, Claude Sonnet 4.5, Gemini 3 Pro) employ an extended chain of thought before each action, trading higher token consumption for improved multi-step planning. Standard code-generation agents (GPT-5.2 Codex, GPT-5.3 Codex, Gemini 2.5 Flash, Gemini 3 Flash, Kimi K2.5 Coding) operate in a direct prompt-to-action mode optimized for code output. Lightweight agents (Claude Haiku 4.5, Qwen3 Coder+) are designed for speed and cost efficiency. All agents manage their own context window internally, employing provider-specific strategies for long-context handling (context pruning, summarisation, or sliding-window attention) that are not configurable by the user. All agents operate in their default fully autonomous mode with no interactive confirmation steps.

All agents interact with the execution environment through a single-process command-line interface with full read-write access to the project directory and a bash shell. The agent reads files from disk (including the staged domain protocols and diagnostic recovery mechanisms), writes Python or shell scripts, and executes them via subprocess calls, observing stdout and stderr directly. Pre-compiled VIC and routing binaries are invoked as shell commands. No API mediation, multi-agent routing, or shared-memory coordination is involved; each agent operates as a single autonomous process that must plan, execute, and debug the entire pipeline through sequential file-system interactions. This architecture is fundamentally distinct from API-mediated multi-agent frameworks in that the agent must discover the directory structure, parse file contents, and manage long-running processes (VIC execution takes 5--20\,minutes) without any scaffolding beyond its native CLI capabilities.

Token consumption is standardized across providers: input tokens exclude cache reads; output tokens include all generated content. Cache read tokens are tracked separately. Cost is computed from each provider's published per-token pricing at the time of evaluation. For the Gemini family, token counts were not reliably captured by the logging harness and are excluded from token analyses; success rates for Gemini models are drawn from a separate set of runs verified to be free of API rate-limiting (HTTP 429) errors.

\subsection{Benchmark design for agents using the VIC-routing knowledge infrastructure}

Each benchmark trial proceeds as follows. A CLI coding agent receives a single natural-language prompt specifying the target basin, simulation period, along with the complete knowledge infrastructure package for VIC-routing (detailed prompt in Supplementary Information Note~2). The agent operates in a fresh containerized Linux instance (Ubuntu 22.04) containing precompiled VIC 5.1.0 and Lohmann routing executables, raw input datasets (meteorological forcing, soil properties, land cover, digital elevation model, basin shapefiles), and observed discharge for evaluation. No human intervention occurs after the initial prompt. The agent must autonomously read staged domain protocols, invoke validated modelling operators, diagnose failures using diagnostic recovery mechanisms, and produce routed discharge at the target gauging station within a 60-minute wall-clock timeout, set at twice the time required for a domain expert to complete the pipeline manually.

We executed 100 independent trials per model--basin combination across 10 agents and 3 basins (3{,}000 trials total). Where extended sampling produced more than 100 trials for a model--basin combination, only the first 100 were used in all analyses, ensuring identical denominators across non-Gemini agents. For the Gemini family, persistent HTTP 429 errors prevented stable accumulation of 100 trials per basin within the original sampling window; success rates and the failure-stage distribution shown in Fig.~4a are therefore taken from a separate 100-per-basin re-test verified to be free of rate limiting. Each agent is stateless across trials, with no conversational memory, cached outputs, or learned strategies carrying over, and each trial accesses a fresh copy of the environment. At $n = 100$ trials per basin, the 95\% Clopper--Pearson confidence interval has a half-width of 7.6 percentage points at the observed Tier 1 rate of 84\%, sufficient to distinguish adjacent models.

We assess success based on the 14 milestones and divide them into five pipeline phases: domain preparation (milestones 1 to 3: basin delineation, grid generation, soil and forcing preparation), VIC configuration and execution (milestones 4 to 6: vegetation processing, global parameter file generation, model execution), output postprocessing (milestones 7 to 9: column extraction, flow direction derivation, routing parameter construction), routing execution (milestones 10 to 12: station file generation, configuration, model execution), and evaluation (milestones 13 to 14: discharge comparison, metric computation). Each milestone has programmatic pass/fail criteria, including checks for file existence, column count validation, coordinate precision verification, and value range constraints. A trial is classified as successful when all 14 milestones pass and a valid Nash--Sutcliffe efficiency is computed against observed discharge. Trials that advance past at least one milestone but do not complete are classified as partial. Trials terminated by the 60-minute timeout or that fail to advance beyond initial setup are classified as failed.

\subsection{Failure classification and ablation experiment design}

Failed trials are classified along two axes. The first identifies the pipeline phase at which the agent stalled, determined by the last milestone successfully reached before session termination; this classification is fully automated from milestone checkpoint logs. The second categorizes the root cause of failure into five types. Two types of infrastructure constraints are reflected: execution timeout (the agent exceeded the 60-minute limit while still actively working, indicated by a SIGTERM exit code of $-15$) and platform instability (API disconnection, rate limiting, or server-side errors, identified from structured error logs). Three types reflect agent behavioural limitations: instruction adherence (the agent deviated from documented parameter specifications or tool calling conventions), repetitive behaviour (the agent entered debugging loops attempting the same failed approach three or more consecutive times), and workflow reasoning (the agent invented undocumented steps, skipped dependencies, or reordered the pipeline). These five categories partition into two higher-level attributions: infrastructure constraints and agent behavioural limitations, distinguishing failures addressable by computational resource improvements from those requiring advances in agent reasoning.

To isolate the contribution of knowledge infrastructure from each agent's intrinsic coding and reasoning capabilities, we conduct a clean-room ablation experiment. Six models are selected to cover all five platforms and include the top-performing agents: Claude Opus and Claude Sonnet (Anthropic), GPT-5.3 Codex (OpenAI), Gemini 3 Flash (Google), Kimi Coding (Moonshot AI), and Qwen3 Coder+ (Alibaba). Each model receives 10 trials on the Wangjiaba basin under bare-model conditions: the same natural language prompt and raw input datasets, but with all knowledge infrastructure removed. No validated modelling operators, no staged domain protocols, no diagnostic recovery mechanisms. The agent must write all processing scripts from scratch, infer the correct pipeline sequence, and debug failures using only its pre-training knowledge and the error messages produced by VIC and the routing model. The 60-minute timeout is maintained.

\subsection{Knowledge dissection for the 119 models}

Dissection was carried out at three levels of automation across the 119-model cohort, each producing KI packages conforming to a common 9-stage pipeline (Fig.~\ref{fig:fig2}a): source acquisition, I/O pipeline mapping, unit discovery, tool generation, SKILL.md, diagnostic-triplet construction, package assembly, binary test, and progressive validation. For VIC-routing, a domain expert executed the complete workflow manually for each study basin, recording every command, intermediate file, and debugging decision. These records were classified into the three knowledge types and encoded into the corresponding layers. Each component was tested through targeted ablation, in which individual tools or protocols were removed, and the resulting agent failure was compared with the predicted failure mode. Three revision cycles over six months produced the final package, which was formalized into a standardized schema defining the required fields for pipeline stages, tools, protocols, and triplets.

For the next 25 models, spanning eight Earth-science domains, construction was delegated to the single-agent Knowledge Dissection Toolkit while validation remained expert-supervised. KDT clones the model source, scans its input/output structure, reads the official documentation, and executes the 9-stage pipeline: stages s0--s1 (source probing and I/O tracing) are fully automated, and stages s2--s8 (unit discovery, tool generation, SKILL.md authoring against a 15-section template, diagnostic-triplet construction, package assembly, binary test, and progressive validation) are agent-guided. The authors reviewed each phase and assessed correctness at validation checkpoints. A PREFLIGHT checklist of 11 trap categories, Fortran, coupling, unit, spatial, domain, platform, lake, flood, config-file, multi-structure, and flow-network traps, was applied before each new model, and both the checklist and a cross-model error-pattern library were updated after each wave. Average dissection time fell from weeks per model in Waves 0--2 to 2--4\,hours in Waves 3--5.

The remaining 92 models were dissected fully autonomously by the multi-agent extension of KDT. Four specialized agents operate in sequence on each model, with a fifth (the Fixer) activated only on retry. The Builder compiles the binary and verifies that it runs. The KI Generator produces the package through the 9-stage pipeline. A fresh Tester agent, which reads only the packaged KI and never the source code, runs the model against an observation site sampled independently from the HydroCraft observation database. The Reviewer cross-checks the package against the shared schema. When the Tester fails, the Fixer receives the failure report, runs five automated diagnostic checks (water-balance closure, forcing-range plausibility, output-unit consistency, parameter consistency, and decision-flag compatibility), and patches the offending tool or SKILL.md entry, and returns control to the Tester, with up to three retry cycles per model. Each completed package was then entered into observation-based validation, with full implementation details provided in Supplementary Method~2.

\subsection{Knowledge infrastructure validation}

Every KI package was subjected to a common staged validation procedure, but the final validation depth depended on model type and data availability. Step 1 confirmed that the model binary compiled and ran against developer-supplied test data. Step 2 progressively replaced input components with HydroCraft-prepared global or regional data to isolate adapter errors. Step 3 required observation-based validation where suitable independent observations were available; otherwise, packages were verified for end-to-end runnability using synthetic, example or analytical inputs shipped with the source model.

The hydrological models in the 25 expert-supervised cohorts were assigned to the same three Huai River basins as the VIC-routing benchmark (Xixian, Wangjiaba, Bengbu), so that simulated discharge could be compared directly with VIC-routing outputs, with observed discharge, and across models. The non-hydrological models in the 25 expert-supervised cohorts (crop, biogeochemistry, ocean, cryosphere, water quality) were given access to the shared HydroCraft observation database (FLUXNET2015, FAOSTAT, NOAA tides and buoys, SNOTEL, USGS sediment, EPA National Lake Assessment and others) and chose their own first validation site from within the domain-appropriate sources; that site was then fixed for runs 2 and 3 to measure run-to-run reproducibility under identical KI. The 92 fully auto-dissected packages were validated end-to-end by the cross-check extension of KDT. In this mode, the agent selected its own validation site for each run from the same observation database, and four cross-checking agents (Runner, Observer, Calculator, Auditor) operated as an independent verification chain. The Runner executed the model using only the packaged KI, the Observer fetched the matching observational record in a parallel, isolated session, the Calculator recomputed performance metrics directly from the raw output and observation files, and the Auditor gated each database write behind an explicit APPROVED/REJECTED decision.

Reported metrics were computed against domain-appropriate observational targets: Pearson $r$ for time-series outputs (discharge, GPP, NEE, latent and sensible heat, surface water elevation, soil temperature at depth), percent bias ($|\text{PBIAS}| \leq 25\%$) for annual or seasonal aggregates (crop yield, sediment load, burn area), and domain-specific metrics for the remaining categories (velocity RMSE for ice sheets, Jaccard index for fire perimeters). After each validation run, the authors compared the reported output with the raw log files to confirm that the model binary had executed and that the metrics reported by the agent matched those in the logs. This post-hoc audit verified only the results; it did not re-examine the agent's decision path.

\subsection{Structural analysis of the 119-model cohort}

To test whether the knowledge-infrastructure (KI) structure is domain-invariant rather than hydrology-specific, we audited the 119-model cohort at three levels: tool-function classification, triplet failure-mode classification, and decision-point extraction, followed by inter-rater reliability and a cross-domain concordance test. For \textbf{tools} and \textbf{failures}, both category schemes are inductively derived from the cohort itself. Tool functions and triplet failure modes were classified using rule-based regular-expression parsers applied to the model code and its KI records, respectively. The decision-point taxonomy was derived in two passes from internal sources within the cohort.

The 75 preflight heuristics, already catalogued from earlier dissections and each attached to a named choice trap such as forcing source, unit system, spatial domain, or parameter choice, provided a starting set of recurring choice axes. We then sampled KI documentation across the 119 models and clustered the configuration decisions we observed along the following axes: families that merged cleanly were kept as single categories; residual clusters that did not match any preflight axis were promoted to new categories; singletons were folded into an OTHER catch-all. The process converged on eleven categories. An agent then read the KI for each model, identified every discrete configuration decision, and emitted a record for each decision containing the model, domain, plain-language description, category label, number of options faced, and documentation source. The agent was given the 11 category names, but no per-category definitions, so category boundaries were tested empirically rather than prescribed, yielding 3478 decisions across the 119 models.

Inter-rater reliability for the decision-point classification was quantified on a stratified sample of $n = 100$ decisions, re-coded independently and in a blind manner by two human auditors. Cohen's $\kappa$ for agent-vs-auditor was $0.81$ and $0.82$; inter-auditor $\kappa$ was $0.96$; Fleiss' $\kappa$ across all three raters was $0.86$ ($\kappa \geq 0.80$). Cross-domain concordance of failure-mode structure was tested by pairwise Spearman rank correlation between each pair of domains' twelve-dimensional failure-category proportion vectors (14 domains, 91 pairs; significance assessed at $\alpha = 0.05$, two-sided). The median $\rho$ was $0.75$, indicating a domain-invariant failure spectrum.

\section{Discussion}

Knowledge infrastructure proved a stronger predictor of reliable simulation than agent capability. The deeper finding is that operational expertise across 14 Earth-science domains converges into a shared structure rather than a domain-specific exception. This makes KI the missing operational layer for what Earth-system science has long called for: integrated modelling across disciplinary communities \citep{Wood2011,Bierkens2015}, achievable not through monolithic frameworks but through a shared scaffold any agent can navigate. Each community can dissect its own models into KI packages; agents compose across them. The breakthrough is therefore not a platform but a protocol that the modelling community can collectively own and extend.

Physics-AI coupling has typically meant either replacing physics with surrogates or embedding physics constraints inside neural networks. KI represents a third mode: agents operate process-based physics models end-to-end, leaving the equations untouched while opening the operational interface. This reframing has both methodological and practical consequences. Methodologically, KI preserves the auditable physics that have made process-based models scientifically credible while making them operationally tractable for general-purpose agents. Practically, the same scaffold that enables intercomparison and cross-domain coupling within the modelling community also lowers the access barrier between process-based simulation and the communities that most need its outputs (Fig.~\ref{fig:fig6}).

Several limitations define the boundary of these conclusions. First, the VIC-routing benchmark estimates completion reliability under controlled conditions, but 100 trials per agent--basin combination do not fully characterize provider-side variability, platform updates or long-term agent drift. Second, the 25 expert-supervised KI packages were validated through agent-driven trials, whereas the 92 auto-dissected packages have not all undergone equivalent domain-expert review. Within the autonomous cohort, observational validation was possible for many models, but others were verified only for end-to-end runnability on synthetic, example or analytical inputs. These packages, therefore, establish agent-operable execution and staged validation, not universal scientific certification for every model or question. Third, the conserved decision and failure structures observed here are demonstrated within Earth-system process-based models. Whether the same scaffold extends to other scientific modelling domains remains an open empirical question. Fourth, KI removes operational ambiguity but does not resolve the structural uncertainties of the underlying models themselves. An agent that runs a model correctly still inherits whatever assumptions, parameterizations, and missing processes that model carries.

These limitations do not weaken the need for KI; they define the community task ahead. KI is not a replacement for domain expertise, calibration, field data or model judgement. It is a way to make more of that expertise explicit, auditable and reusable. Each KI package functions as a standardised interface to a single model. It encapsulates the operational expertise required to run that model correctly: which inputs need which conversions, which configurations apply to which physical regimes, and which intermediate diagnostics separate a physically valid simulation from a plausible-looking failure. Once captured this way, that expertise becomes portable. Any compatible agent can load the package, and any researcher who reads it can audit, extend or correct what it encodes. Earth-system science already has validated models, open-source codes and expert communities that know how to test and trust them. What has been missing is a shared scaffold through which this knowledge can become usable by agents and accessible beyond its original community. The value of KI will therefore depend on collective participation. Hydrologists, crop modellers, biogeochemists, cryosphere scientists, hazard researchers, and others must contribute, validate, maintain and extend KI for their own models. We release KDT and HydroCraft as a starting point, with the expectation that each community will build, audit and version-control its own packages over time. The advance is not a single platform or a single agent, but a community-built interface that makes process-based models more inspectable, interoperable and accessible across science. By making operational expertise re-executable rather than tacit, KI changes who can use these models and how researchers can work across them.

\section*{Data and code availability}

All datasets used in this study are listed in Supplementary Table~3, including the provider, access URL, and primary reference for each; the most relied-upon datasets are CMFD (\url{https://data.tpdc.ac.cn}) and HWSD (\url{https://www.fao.org/soils-portal}). The VIC 5.1.0 source code is available at \url{https://github.com/UW-Hydro/VIC}. The Knowledge Dissection Toolkit is deposited at \url{https://github.com/lzwei196/knowledge_dissection_toolkit}, and a platform for running all 119 models with KI is available at \url{https://app.hydrocraft.ai}. Discharge records from the Chinese hydrological yearbooks for Xixian, Wangjiaba and Bengbu are subject to data-sharing agreements with the relevant Chinese hydrological authorities and are available from the corresponding author upon reasonable request.

\begin{ack}
We gratefully acknowledge funding provided by the National Natural Science Foundation of China (52279018, 52121006, and U2240203), National Basic Research Special Fund for Public Welfare Research Institutes (YDS25002), 2025 Innovation Fund of the National Key Laboratory of Water Disaster Prevention (Yk325002, 524015252, 5240152M2), Fundamental Research Funds for the Central Universities (B240201060), Open Research Fund of National Engineering Research Center of Water Resources Efficient Utilization and Engineering Safety (1524020003).
\end{ack}

\section*{Author contributions}

Z.L. designed the study. Z.L., L.Z. and B.L. led the knowledge dissection of the 119-model cohort across hydrology, agriculture, biogeochemistry, cryosphere and water-quality domains. Y.Z. designed and ran the agent benchmark and ablation experiments, with support from Y.L. on agent integration. Z.L. analysed the failure-mode, decision-point and reproducibility data with input from L.Z. Z.L. and Y.L. wrote the first draft of the manuscript with input from L.Z., B.L. and J.J. All authors contributed to the interpretation of results and approved the final manuscript. J.J. and J.Z. supervised the work and acquired funding.

\section*{Competing interests}

The authors declare no competing interests.


\clearpage
\section*{Appendix A: Extended Data}
\addcontentsline{toc}{section}{Appendix A: Extended Data}

\begin{figure}[!htbp]
  \centering
  \includegraphics[width=0.95\linewidth]{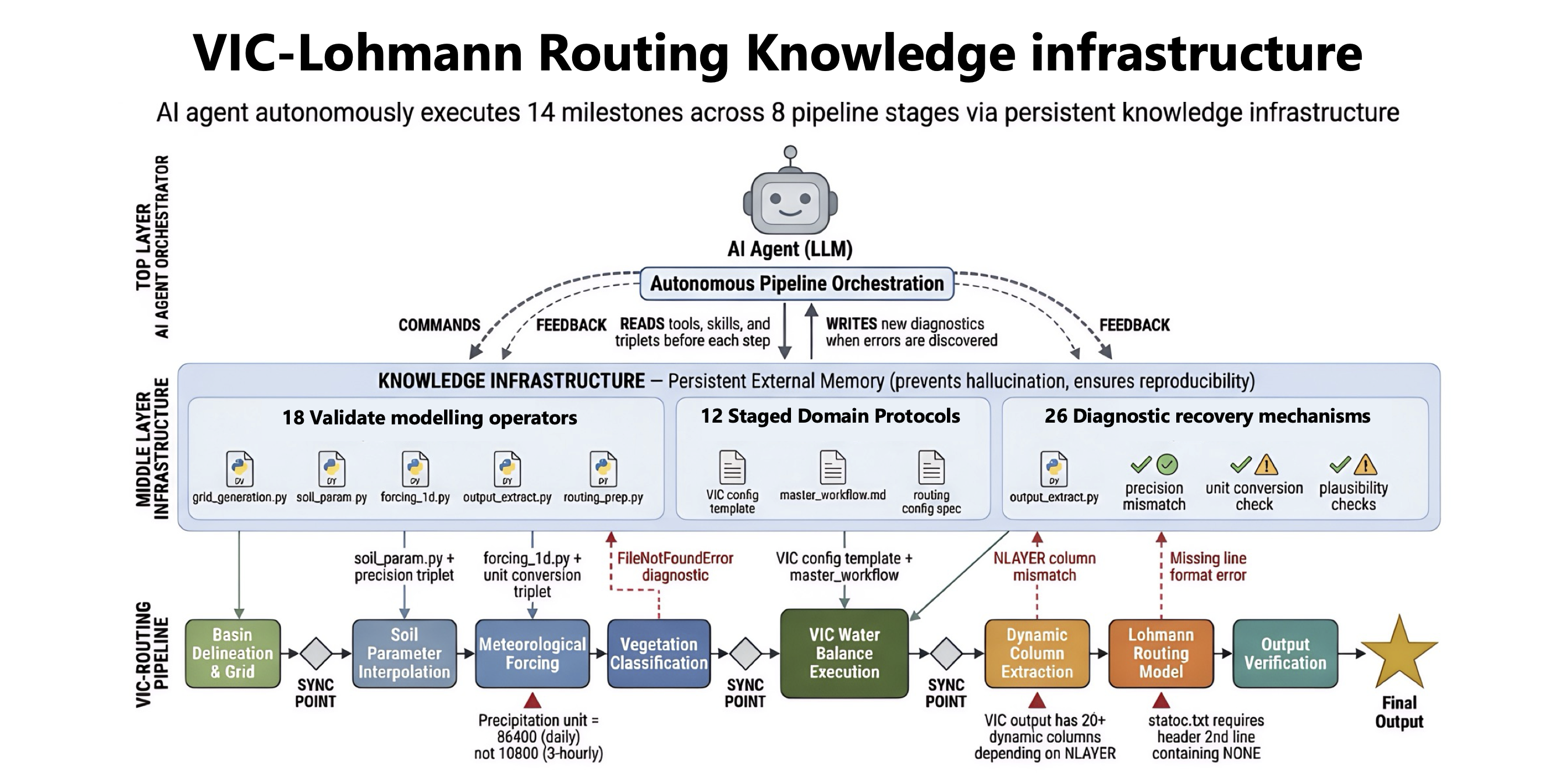}
  \caption{\textbf{Extended Data Fig.~1} Knowledge-augmented pipeline architecture for VIC-routing. The top layer shows the autonomous pipeline orchestration, the middle layer contains the three knowledge infrastructure components, and the bottom layer shows the VIC-routing pipeline from basin delineation through to final output verification. Sync points enforce cross-branch consistency at critical junctures.}
  \label{fig:edfig1}
\end{figure}

\begin{figure}[!htbp]
  \centering
  \includegraphics[width=0.85\linewidth]{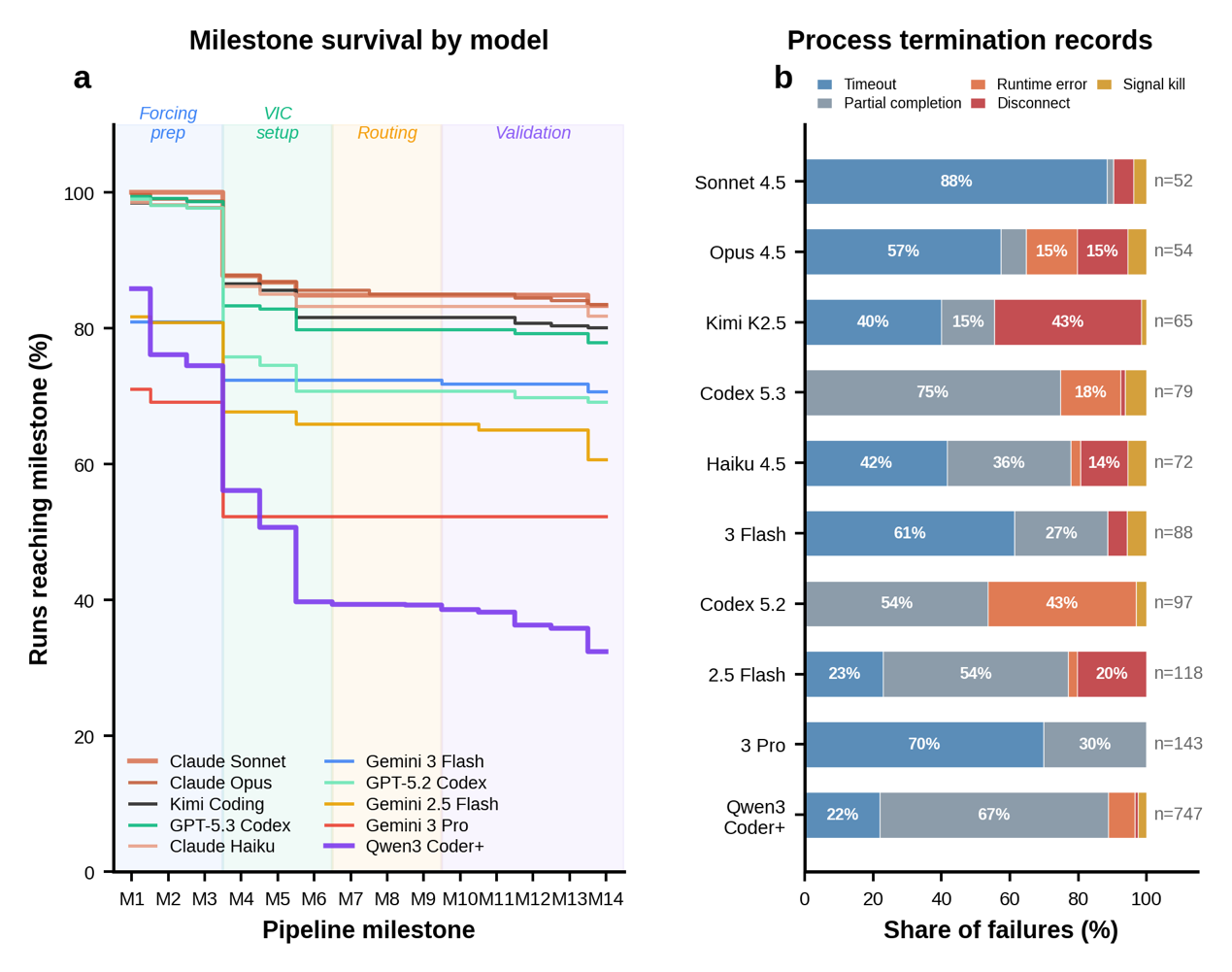}
  \caption{\textbf{Extended Data Fig.~2 Milestone survival and process termination.} \textbf{(a):} Milestone survival curves for all 10 models. Step functions show the percentage of runs reaching each pipeline milestone (M1--M14). Phase annotations indicate the four pipeline stages: forcing preparation (M1--M3), VIC setup (M4--M6), routing (M7--M9), execution (M10--M12), and evaluation (M13--M14). Tier 1 agents maintain near-complete survival through all 14 milestones, while lower-tier agents show steep attrition before M4. \textbf{(b):} Process termination records by model. Stacked bars show the proportion of failures classified by exit code: timeout (SIGTERM), partial completion (exit 0), runtime error (exit 1), platform disconnect (exit $-1$), and signal kill (other signals). Per-model signatures vary: Claude Sonnet failures were predominantly timeouts (88\%), while Qwen3 Coder+ failures were dominated by partial completions (67\%).}
  \label{fig:edfig2}
\end{figure}

\begin{figure}[!htbp]
  \centering
  \includegraphics[width=0.80\linewidth]{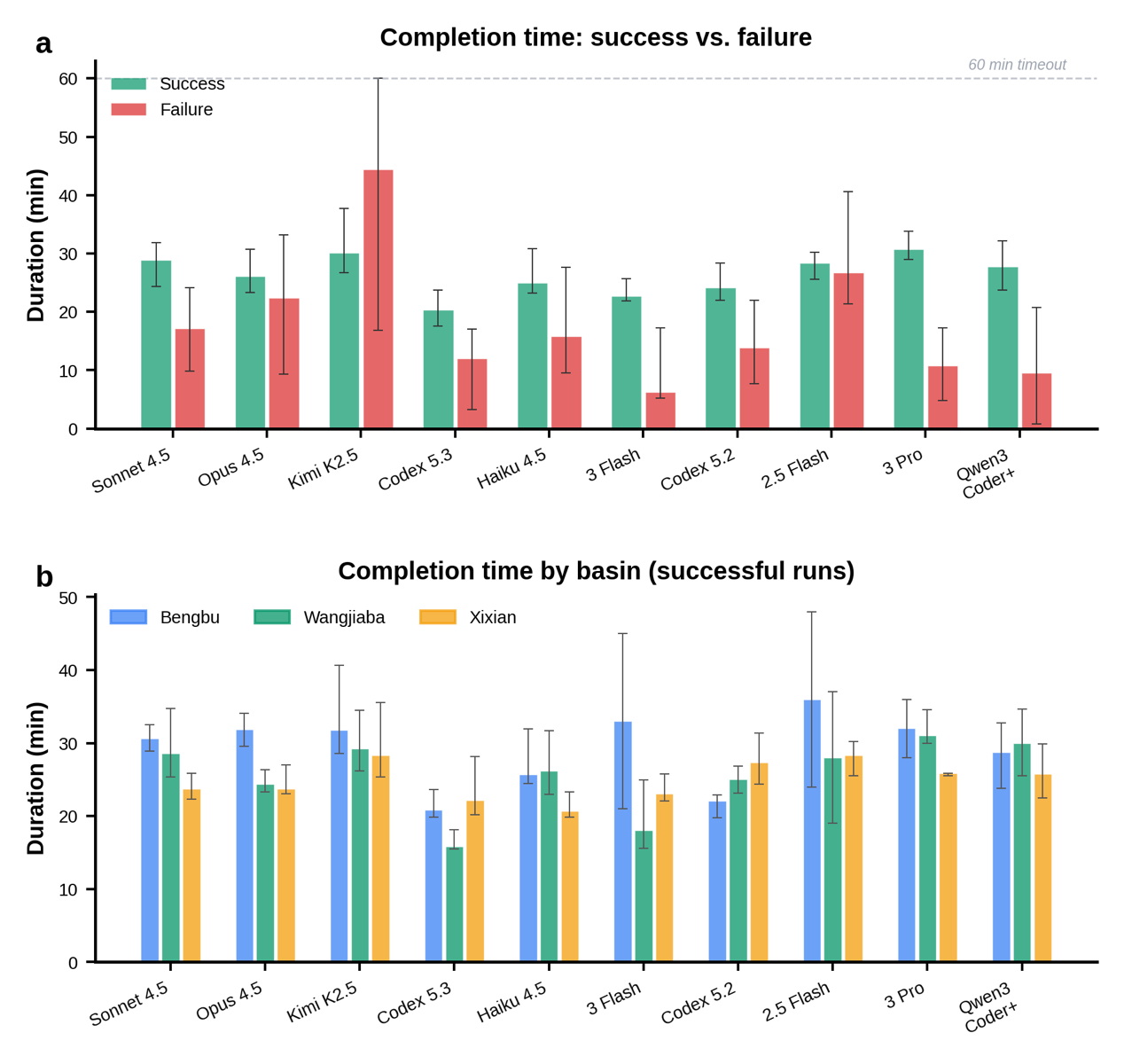}
  \caption{\textbf{Extended Data Fig.~3 Completion time analysis.} \textbf{(a):} Completion time for successful versus failed runs per model. Successful runs cluster between 20 and 35\,minutes, while failed runs show higher variance. \textbf{(b):} Completion time by basin for successful runs.}
  \label{fig:edfig3}
\end{figure}

\begin{figure}[!htbp]
  \centering
  \includegraphics[width=0.85\linewidth]{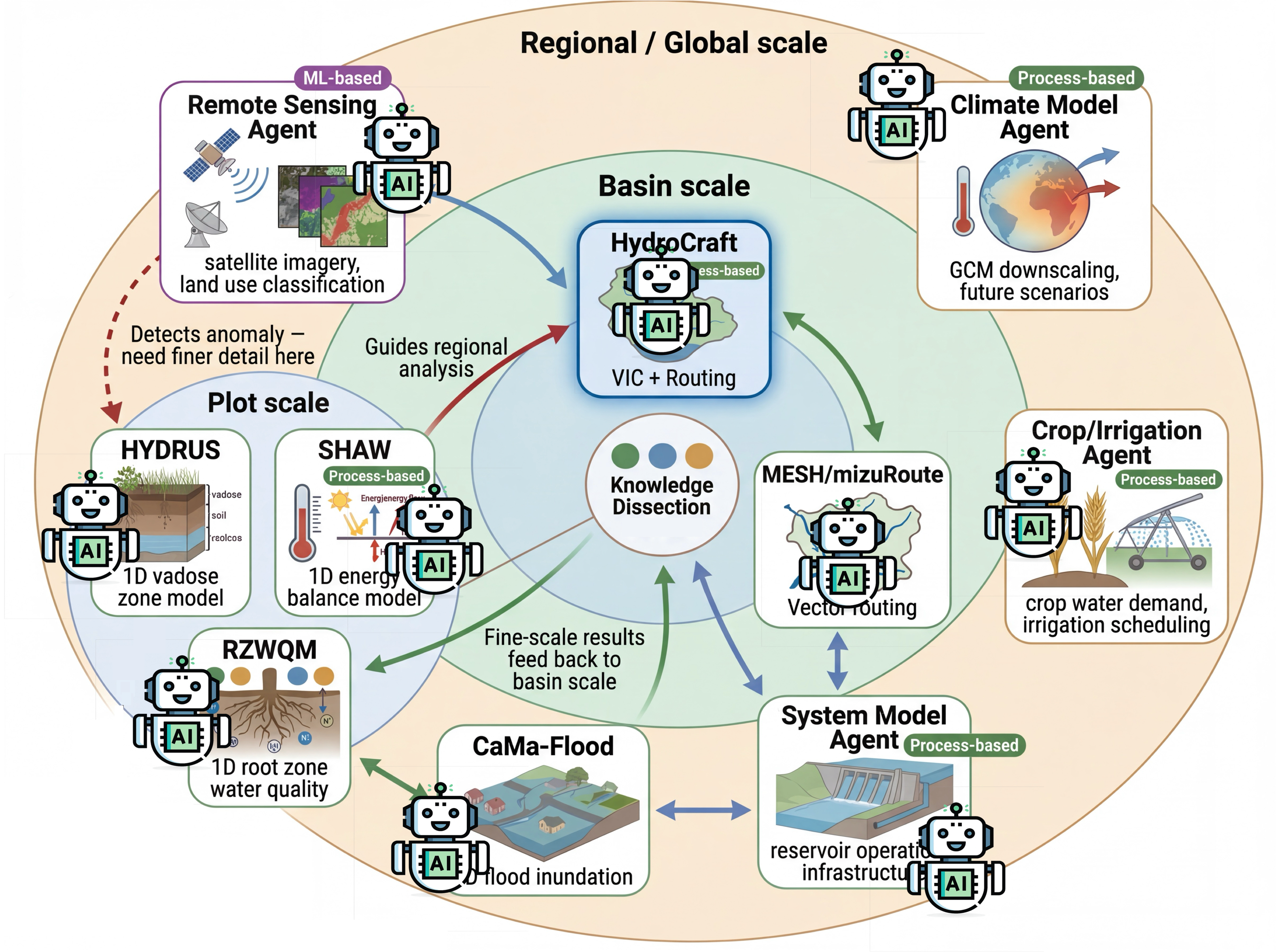}
  \caption{\textbf{Extended Data Fig.~4 Knowledge infrastructure as a scalable pattern for multi-agent Earth system modelling and expanded scientific access.} Conceptual architecture for dynamic, multi-scale simulation. Each model agent wraps a process-based or machine-learning model with its dissected knowledge infrastructure (validated modelling operators, staged domain protocols, diagnostic recovery mechanisms) and communicates with other agents through standardized interfaces. Agents operate at three spatial scales: plot-level models (HYDRUS, SHAW, RZWQM2) capture fine-resolution soil and energy balance processes; basin-scale agents (HydroCraft for VIC-routing, MESH/mizuRoute, Crop/Irrigation) simulate hydrology, routing, and agricultural water demand; and regional agents (Remote Sensing, Climate Model) provide satellite-derived boundary conditions and GCM-downscaled forcing. Knowledge dissection (centre) is the reusable core that enables each agent to be constructed independently and coupled through negotiation of shared inputs and outputs, reducing integration complexity from $O(N^2)$ bespoke interfaces to $O(N)$ knowledge infrastructure packages.}
  \label{fig:edfig4}
\end{figure}

\begin{figure}[!htbp]
  \centering
  \includegraphics[width=0.69\linewidth]{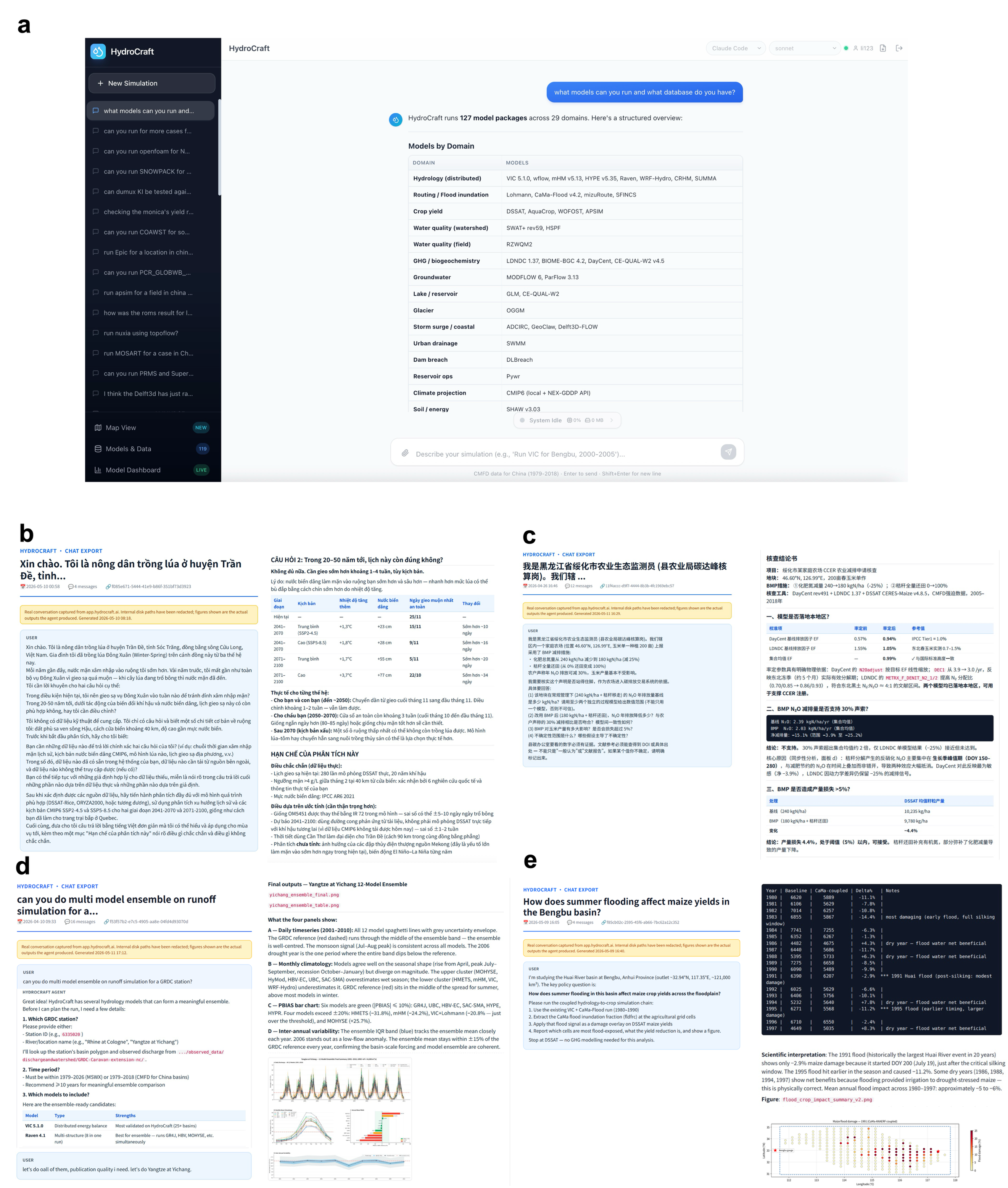}
  \caption{\small \textbf{Extended Data Fig.~5.} \textbf{(a): HydroCraft platform interface and live model catalogue.} Screenshot of the deployed HydroCraft platform (app.hydrocraft.ai) showing the natural-language chat interface (right) and the active model catalogue browser (center table). Users describe a simulation in plain language, and the platform dispatches a CLI coding agent equipped with the relevant KI package to execute it end-to-end. The snapshot shown here lists 127 model packages, more than the 119 reported in the main text. This difference reflects the fact that the platform runs the multi-agent KDT continuously and autonomously on newly identified model candidates, so the catalogue grows between manuscript snapshots, and the live count at app.hydrocraft.ai at any subsequent visit may be higher than the number reported here. The 29-domain grouping shown in the side menu is a finer browsing taxonomy used for catalogue navigation; the structural analysis in the main text aggregates these into 14 Earth-science domains. \textbf{(b--e):} Verbatim chat exports of the four proof-of-concept demonstrations summarised in Fig.~\ref{fig:fig6}b--e. Each panel reproduces the user's natural-language question (left, highlighted), the agent's textual reasoning and intermediate outputs, and the headline result (right). No translation, paraphrasing, or post-hoc editing has been applied to any panel; the full session transcripts are available via the platform. \textbf{(b):} Mekong Delta rice farmer (Fig.~\ref{fig:fig6}b). A Vietnamese-speaking smallholder in Tr\`{a}n {\DJ}\^{e} district, S\'{o}c Tr\u{a}ng asks about salinity-sensitive sowing windows. The agent loads the AquaCrop KI, runs the simulation against local CMFD forcing, and returns salinity-aware rice yield projections directionally consistent with IRRI CS-MAP recommendations \citep{Ferrer2022}. \textbf{(c):} Suihua MRV officer (Fig.~\ref{fig:fig6}c). A county-level Chinese-language carbon-credit screening request for a CCER claim. The agent dispatches a three-model ensemble (DSSAT + LDNDC + DayCent), surfaces a 55-percentage-point inter-model divergence, diagnoses two configuration errors, calibrates to local NE China black-soil parameters, and returns a conservative ensemble verdict flagging the claim's 30\% N\textsubscript{2}O reduction as unsupported. \textbf{(d):} Yangtze multi-model runoff ensemble (Fig.~\ref{fig:fig6}d). The user requests a multi-model runoff ensemble at a Yangtze station. The agent assembles twelve runoff models across six modelling families in a single session, returns aligned hydrographs, and produces inter-model spread diagnostics, a workflow that conventionally requires multi-PI coordination across modelling groups. \textbf{(e):} Bengbu cross-domain coupling (Fig.~\ref{fig:fig6}e). A user asks how summer flooding affects maize yields in the Huai River basin near Bengbu. The agent constructs a VIC + CaMa-Flood + DSSAT-Maize coupling chain, runs runoff and inundation simulations, and crop-impact simulations, and returns a linked process chain through a single inspectable workflow.}
  \label{fig:edfig5}
\end{figure}

\clearpage

\begin{table}[!htbp]
  \caption{\textbf{Extended Data Table 1.} CLI coding agent specifications.}
  \label{tab:agents}
  \centering
  \small
  \begin{tabular}{p{2.2cm}p{1.6cm}p{2.4cm}p{1.5cm}p{1.5cm}p{1.2cm}p{1.6cm}p{0.8cm}}
    \toprule
    Agent variant & Platform & Foundation model & Parameters (active / total) & Architecture\textsuperscript{b} & Context window\textsuperscript{c} & Reasoning mode & Open weights \\
    \midrule
    Claude Sonnet 4.5 & Claude Code & claude-sonnet-4-5-20250929 & Not disclosed\textsuperscript{a} & Not disclosed & 200{,}000 & Extended thinking & No \\
    Claude Opus 4.5 & Claude Code & claude-opus-4-5-20251101 & Not disclosed & Not disclosed & 200{,}000 & Extended thinking & No \\
    Claude Haiku 4.5 & Claude Code & claude-haiku-4-5-20251001 & Not disclosed & Not disclosed & 200{,}000 & Extended thinking & No \\
    GPT-5.2 Codex & Codex CLI & codex-5.2 & Not disclosed & Not disclosed & 400{,}000 & Adjustable (low/medium/high) & No \\
    GPT-5.3 Codex & Codex CLI & codex-5.3 & Not disclosed & Not disclosed & 400{,}000 & Adjustable (low/medium/high) & No \\
    Gemini 2.5 Flash & Gemini CLI & gemini-2.5-flash & Not disclosed & Not disclosed & 1{,}000{,}000 & Configurable thinking budget & No \\
    Gemini 3 Flash & Gemini CLI & gemini-3.0-flash & Not disclosed & Not disclosed & 1{,}000{,}000 & Adjustable thinking levels & No \\
    Gemini 3 Pro & Gemini CLI & gemini-3.0-pro & Not disclosed & Not disclosed & 1{,}000{,}000 & Adjustable thinking levels & No \\
    Kimi K2.5 Coding & Kimi Code CLI & kimi-k2.5 & 32B / 1T & Mixture-of-Experts & 256{,}000 & Chain-of-thought & Yes \\
    Qwen3 Coder+ & Qwen CLI & qwen3-coder-plus & 35B / 480B & Mixture-of-Experts & 256{,}000 & Hybrid thinking & Yes \\
    \bottomrule
  \end{tabular}

  \vspace{0.5em}
  \noindent\textbf{Notes:} All reasoning modes were left at platform defaults during evaluation; no manual tuning of thinking levels or reasoning effort was applied. \textsuperscript{a}: ``Not disclosed'' indicates the provider has not publicly released the specification as of March 2026. \textsuperscript{b}: Architecture classifications are based on publicly available information. Mixture-of-Experts (MoE) routes each token through a subset of total parameters; providers that have not disclosed their architecture are listed as ``Not disclosed'' rather than assumed. \textsuperscript{c}: Context windows reflect the advertised maximum at the time of evaluation; effective usable context is smaller due to system prompt overhead (e.g., $\sim 155{,}000$--$167{,}000$ usable tokens for Claude Code).
\end{table}

\begin{table}[!htbp]
  \caption{\textbf{Extended Data Table 2.} AI-guided calibration benchmark.}
  \label{tab:calibration}
  \centering
  \small
  \begin{tabular}{lccccccc}
    \toprule
    Model & Success rate ($\geq 0.7$) & Mean NSE & Best NSE & NSE S.D. & Mean iter. & Mean dur. (min) \\
    \midrule
    Claude Opus 4.5 & 10/10 (100\%) & 0.706 & 0.729 & 0.008 & 34.1 & 128 \\
    Codex 5.3       & 10/10 (100\%) & 0.703 & 0.707 & 0.003 & 21.5 & 74  \\
    Codex 5.2       & 7/10 (70\%)   & 0.697 & 0.719 & 0.027 & 20.8 & 70  \\
    Gemini 3 Flash  & 5/10 (50\%)   & 0.697 & 0.716 & 0.011 & 40.8 & 135 \\
    Claude Sonnet 4.5 & 4/10 (40\%) & 0.675 & 0.704 & 0.029 & 34.4 & 117 \\
    Kimi K2.5       & 1/10 (10\%)   & 0.602 & 0.708 & 0.056 & 3.9  & 19  \\
    \bottomrule
  \end{tabular}

  \vspace{0.5em}
  \noindent\textbf{Note:} Six AI models attempted iterative calibration of the VIC hydrological model on the Bengbu basin (1980--1990), optimizing six parameters (\texttt{binfilt}, \texttt{Ds}, \texttt{Dsmax}, \texttt{Ws}, \texttt{soil\_d2}, \texttt{soil\_d3}) from a baseline NSE of 0.148 toward a target of NSE $\geq 0.7$. Each model performed 10 independent runs with a maximum of 50 iterations per run. Success rate: proportion of runs reaching the target. Mean NSE and Best NSE are computed over all valid runs (including those that did not reach the target). The mean duration is the wall-clock time per run.
\end{table}


\clearpage
\medskip

\bibliography{biblio}

\end{document}